\documentclass[journal]{IEEEtran}
\usepackage{booktabs}
\usepackage[scaled=1.0]{helvet}
\usepackage{times}
\usepackage{graphicx}
\usepackage{subfigure}
\usepackage{parskip}
\usepackage{multirow}
\usepackage{amsmath}
\usepackage{color}
\usepackage[labelfont=bf,textfont=it]{caption}
\usepackage[lined,boxed,commentsnumbered, ruled]{algorithm2e}

\begin{document}
	\title{Unsupervised Learning Based Multi-Scale Exposure Fusion}
	\author{
		C. B. Zheng,~\IEEEmembership{Member~IEEE}, S. Q. Wu,~\IEEEmembership{Senior Member~IEEE}, and Z. G. Li,~\IEEEmembership{Fellow,~IEEE}
		
		\thanks{Chaobing Zheng and Shiqian Wu are with the school of Information Science and Engineering, Wuhan University of Science and Technology, Wuhan 430081, China (email: \{zhengchaobing, shiqian.wu\}@wust.edu.cn).
			Zhengguo Li is with VI department, the Institute for Infocomm Research, A*STAR, Singapore, 138632, (email: ezgli@i2r.a-star.edu.sg).
		}
	}


	\markboth{}%Journal of \LaTeX\ Class Files}%,~Vol.~6, No.~1, January~2007}%
	{Shell \MakeLowercase{\textit{et al.}}: Bare Demo of IEEEtran.cls
		for Journals}



	\maketitle	
\begin{abstract}
	Unsupervised learning based multi-scale exposure fusion (ULMEF) is efficient for fusing differently exposed low dynamic range (LDR) images into a higher quality LDR image for a high dynamic range (HDR) scene. Unlike supervised learning, loss functions play a crucial role in the ULMEF. In this paper, novel loss functions are proposed for the ULMEF and they are defined by using all the images to be fused and other differently exposed images from the same HDR scene. The proposed loss functions can guide the proposed ULMEF to learn more reliable information from the HDR scene than existing loss functions which are defined by only using the set of images to be fused. As such, the quality of the fused image is significantly improved.
	The proposed ULMEF also adopts a multi-scale strategy that includes a multi-scale attention module to effectively preserve the scene depth and local contrast in the fused image. Meanwhile, the proposed ULMEF can be adopted to achieve exposure interpolation and exposure extrapolation. Extensive experiments show that the proposed ULMEF algorithm outperforms state-of-the-art exposure fusion algorithms.
\end{abstract}

\section{Introduction}
\label{sec:intro}
A high contrast nature scene could have a high dynamic range (HDR), with brightness ranging from $10^{-4} cd/m^2$ to $10^6 cd/m^2$, and a dynamic range of up to 10 orders of magnitude. However, the dynamic range that can be captured with a single exposure is very limited, and the recording of image data usually uses 8 bits, resulting low dynamic range (LDR) images which inevitably have unfavorable over-$/$under-exposed regions. Therefore, in extremely bright or dark situations, there will be a significant loss of detailed information, which severely affects machine vision tasks such as intelligent driving and navigation. HDR imaging has been introduced to address the issue effectively \cite{Debevec97}. Due to limited information captured by a single image, single-image based HDR methods usually show poor performances. Capturing multiple differently exposed images provides an efficient solution for HDR imaging, which can preserve rich details and vivid color. Even though camera movements and moving objects are issues for the multiple images, differently exposed LDR images can be aligned in LDR domain by using the algorithm in \cite{1liuz2023}. In the remaining part of this paper, differently exposed LDR images from an HDR scene are assumed to be aligned well as in \cite{1mertens2007,Li2017,kou2017,kou2018,1jia2022,MEFNet,jiang2023meflut,1farbman2008,1vinker2021}.

There are two different ways to combine a set of differently exposed LDR images together for an HDR scene after they are aligned. One is to estimate the camera response functions (CRFs), convert the LDR images into the corresponding HDR images, and merge all the HDR images into one high quality HDR image \cite{Debevec97}. The HDR image is scaled down by a tone mapping algorithm for display \cite{1farbman2008,1vinker2021}. The other is to fuse all the differently exposed LDR images into a high quality LDR image directly by using an exposure fusion algorithm. 

The exposure fusion was widely studied by using conventional methods in \cite{1mertens2007,Li2017,kou2017,kou2018,1jia2022} and data-driven methods in \cite{MEFNet,jiang2023meflut}. The fused image approaches the set of images to be fused rather than the HDR scene by these algorithms. All these algorithms assume that enough differently exposed LDR images are captured with a normal-exposure-ratio (NER) for each HDR scene. However, this assumption is usually not true, especially for mobile devices with limited computational resources. Generally, only a few differently exposed LDR images are captured for an HDR scene. All the exposure fusion algorithms in  \cite{1mertens2007,Li2017,kou2017,kou2018,1jia2022,MEFNet,jiang2023meflut} produce serious brightness order reversal artifacts if the inputs are two large-exposure-ratio (LER) images \cite{1prabhakar2017,1yang2018, zheng2023efficient}. Information in the brightest and darkest regions of an HDR scene might not be preserved well if the inputs are three NER images. Therefore, it is still desired to study the exposure fusion even though there are many exposure fusion algorithms. We argue that the fused image should approach the HDR scene rather than the set of images to be fused. The objective of this paper is to explore such a new exposure fusion algorithm by fully utilizing the asymmetry between the training and inferring (or testing) stages of the learning based algorithm.

Inspired by the algorithms in \cite{1mertens2007,1farbman2008,1prabhakar2017, zheng2023efficient}, a novel unsupervised learning based multi-scale exposure fusion (ULMEF) algorithm is proposed for a set of differently exposed LDR images from an HDR scene in this paper. All the inputs are already aligned. The proposed algorithm is based on an observation that multi-scale is helpful to preserve scene depth and increase detail clarity for the fused LDR image \cite{1farbman2008,1mertens2007}. This is different from the existing unsupervised learning based algorithms in \cite{1prabhakar2017,MEFNet,jiang2023meflut} which are single-scale. Particularly, a multi-scale fusion network (MSF-Net) is proposed  to fuse all the images in the feature domain from coarse to fine. Each scale of the proposed network is on top of multi-scale attention mechanisms which utilize different scale features to fuse the images efficiently, thereby improving the training efficiency of the network. Besides the network structure, loss functions are crucial for the proposed ULMEF algorithm. A new strategy is proposed for the definition of loss functions. The loss functions in \cite{MEFNet,jiang2023meflut, 1prabhakar2017,zheng2023efficient} are tightly coupled with the set of images to be fused. The relative brightness order might not be preserved well if the inputs are two LER images \cite{1yang2018,1zheng2023}, and information in the brightest and darkest regions might not be well preserved in the fused image if the inputs are three NER images. To address these problems, the loss functions are defined by using the set of LDR images to be fused and other differently exposed LDR images from the same HDR scene. As such, the fused image approaches the HDR scene rather than the set of images to be fused. Clearly, the proposed loss functions are fundamentally  different from those in \cite{ MEFNet, jiang2023meflut, 1prabhakar2017, zheng2023efficient} in the sense that the loss functions and the set of images to be fused are decoupled in the proposed ULMEF algorithm. To our best knowledge, we are the first to propose the decoupled loss functions for the unsupervised learning based exposure fusion. The proposed ULMEF can learn more reliable information from the HDR scene than the existing loss functions which are defined by only using the set of images to be fused \cite{MEFNet,jiang2023meflut, 1prabhakar2017,zheng2023efficient}. This is not surprised due to the conventional wisdom of inferring better through seeing more. It can be adopted to achieve exposure interpolation and exposure extrapolation much easier than the conventional MEF algorithms. Experiments on different datasets have demonstrated efficiency of the proposed algorithm. Overall, two main contributions of this paper are

1) An innovative strategy is proposed to define loss functions for unsupervised learning based exposure fusion algorithms. The loss functions and the set of images to be fused are decoupled by the new strategy. As such, the exposure interpolation and exposure extrapolation can be implemented easily. This is a new initiative on exposure fusion. The fused image approaches the HDR scene rather than the set of images to be fused;

2) A novel MSF-Net with multi-scale attention mechanisms is proposed to preserve the scene depth and local contrast in the fused image. In addition, the information in the brightest and darkest regions are preserved and the halos artifacts are avoided from appearing in the fused image by the proposed ULMEF algorithm.

The rest of this paper is organized as below. Existing results on exposure fusion are summarized in Section \ref{literature}. Details of the proposed MEF algorithm are provided in Section \ref{paradigm}. Experiment results are presented in Section \ref{experimentalresult} to compare the proposed algorithm with nine state-of-the-art (SOTA) exposure fusion algorithms. Finally, conclusion remarks are given in Section \ref{conclusion}.

\section{Literature Review on Exposure Fusion}
\label{literature}

Many exposure fusion algorithms were proposed for the HDR imaging under an assumption that all the images to be fused are aligned well. The main idea of these algorithms is to preserve the reliable information from a set of differently exposed LDR images as much as possible. Existing exposure fusion algorithms can be divided into traditional exposure fusion  algorithms and data-driven ones.

Traditional exposure fusion algorithms are mainly based on statistical modeling methods, which perform weighted average or weighted sum of image pixels in a multi-scale way. The resultant algorithm is thus called multi-scale exposure fusion (MEF). Mertens et al. \cite{1mertens2007} first used contrast, saturation, and exposure to define weights for all pixels and then fused the different exposure images to create an information-enriched LDR image by using the Gaussian and Laplacian pyramids \cite{1Burt1983}. This approach allowed for a wider range of brightness and color information to be captured in the final image, resulting in a more realistic and visually appealing representation of the scene. However, the algorithm in \cite{1mertens2007} has a fundamental difficulty in preserving information in the brightest and darkest regions of HDR scenes \cite{1hessel2020}. To address this issue, edge-preserving smoothing (EPS) pyramids and content adaptive edge-preserving smoothing (CAS) pyramids were proposed in \cite{Li2017,kou2017,kou2018,1jia2022}. Since the EPS and CAS pyramids can smoothen the weights, the levels of the pyramids can be reduced. As such, the information in the brightest and darkest regions can be preserved well \cite{1hessel2020}. However, halo artifacts could be an issue for the algorithms in \cite{Li2017,kou2017,kou2018,1jia2022} as indicated in \cite{1hessel2020}. The information in the brightest and darkest regions can also be preserved well by synthesizing more differently exposed LDR images \cite{1hessel2020}.   Many existing MEF algorithms were evaluated and compared in \cite{Cai2018} by using the MEF-SSIM \cite{1Ma2015}. Brightness order reversal artifacts are an issue for all the MEF algorithms in \cite{1mertens2007,Li2017,kou2017,1jia2022} when two LER images are fused by them. Exposure interpolation could be used to avoid the brightness order reversal artifacts from appearing in the fused images \cite{1yang2018,1zheng2023}. Both the halo artifacts and brightness order reversal artifacts will be addressed by the proposed ULMEF algorithm.
It is worth noting that guided filtering for up-sampling (GFU) \cite{1he2013} was extended by using the upsampling methods in the Gaussian and Laplacian pyramids \cite{1Burt1983} to replace the bilinear upsampling in \cite{1he2013}, and the extended GFU was applied to simplify the MEF algorithm in \cite{Li2017}. One beauty of the GFU is that the coefficients of weighted guided image filter (WGIF) \cite{1lizg2015} are only computed at two levels of the pyramids and they are up-sampled to obtain the coefficients of the WGIF at other levels. The other is that the weight maps can be computed from the luminance components and the coefficients of the WGIF at all the other levels. The GFU was adopted by the unsupervised learning based single-scale exposure fusion (ULSEF) algorithms in  \cite{MEFNet,jiang2023meflut} to reduce their complexity.

\begin{figure*}[htb!]
	\centering
	\includegraphics[width=1.0\textwidth]{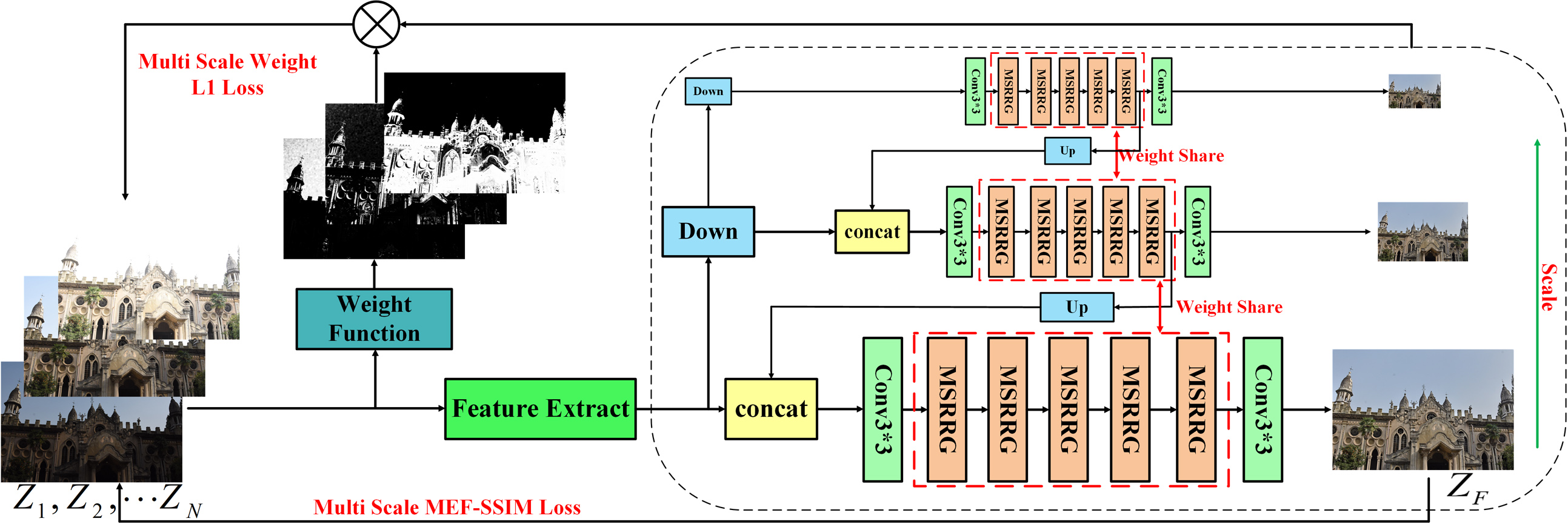}
	\caption{Structure of the proposed MSF-NET. The MSF-NET is on top of a hierarchical structure with three level which is helpful to preserve scene depth and local contrast in a fused image and also improves the MEF-SSIM of the fused image.}
	\label{Fig1}
\end{figure*}

\begin{figure*}[htb]
	\centering
	\includegraphics[width=1.0\textwidth]{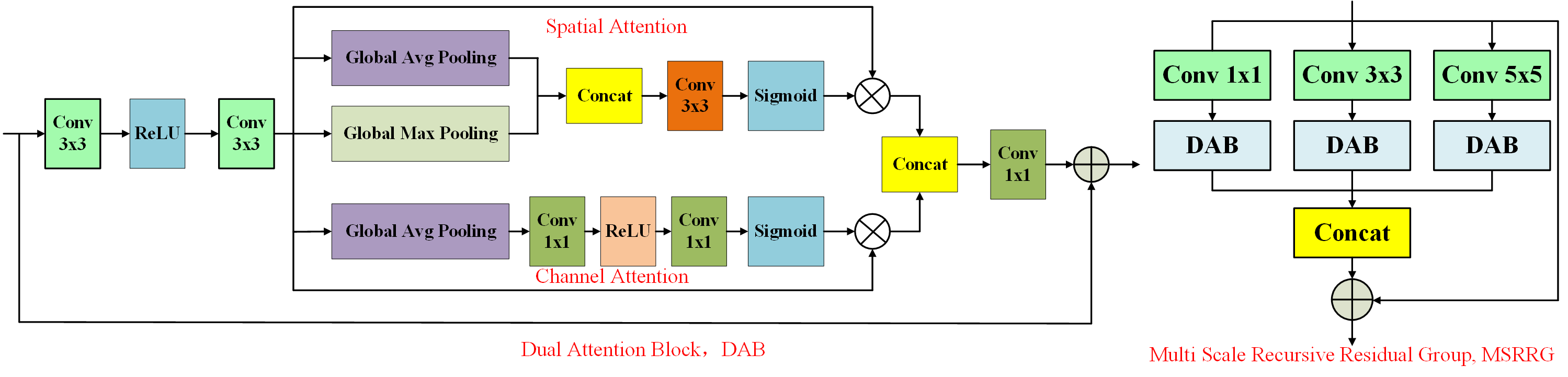}
	\caption{ Multi Scale Recursive Residual Group (MSRRG), each MSRRG contains multi scale dual attention blocks (DAB). Each DAB contains spatial and channel attention modules.}
	\label{Fig2}
\end{figure*}

Confronted with the limited paired training data, most data-driven exposure fusion algorithms are based on un-supervised learning.  The first ULSEF algorithm named DeepFuse \cite{1prabhakar2017} reconstructed an information enriched LDR image from two LER images in YUV color space by using the MEF-SSIM in \cite{1Ma2015}. One more unsupervised ULSEF algorithm for two LER images was proposed in \cite{1xu2020}. Yin et al. \cite{ICME2020} introduced a content prior and a detail prior as guidelines to an encoder-decoder network for two LER images. Prabhakar et al. \cite{cvpr2021} proposed a few-shot learning method to generate labeled dynamic training data from unlabeled one, which greatly released the dependency on labelled ground truth. To release the restrictions on image resolution and exposure number, Ma et al. \cite{MEFNet} introduced an interesting ULSEF algorithm entitled MEF-Net by using the second beauty of the GFU \cite{1he2013}. All the weight maps are first learnt from down-sampled images via unsupervised learning on top of the MEF-SSIM \cite{1Ma2015}, and are then upsampled to the full size via the GFU. Recently, Jiang et al. \cite{jiang2023meflut} proposed a novel and fast ULSEF algorithm by using 1-D look-up tables which are learnt for each exposure by using the unsupervised learning and GFU. The coefficients of the GIFs \cite{1he2013} are computed once by the ULSEF algorithms in \cite{MEFNet,jiang2023meflut}. This is different from the extended GFU in \cite{Li2017}. The coefficients of the WGIF are computed twice in \cite{Li2017} such that the structures of the luminance components are transferred to the weighted maps better at the different levels of the EPS pyramids. This implies that the structures of the luminance components might not be transferred to the weighted maps well by the GFU methods in \cite{MEFNet,jiang2023meflut}, especially when the coefficients of the GIFs are up-sampled by the bilinear upsampling too many times. Subsequently, their performance could be effected. One more issue with the ULSEF algorithms in \cite{MEFNet,jiang2023meflut} is that there are halos in the fused images when the coefficients of the filter are up-sampled too few times for those inputs with small sizes. There are also GAN-based exposure fusion algorithms such as MEF-GAN \cite{mefgan}, etc.
The scene depth and local contrast might not be well preserved by these ULSEF algorithms. The loss function was defined by using the set of images to be fused and the fused image in the ULSEF algorithms \cite{MEFNet,jiang2023meflut}. As such, the fused image approaches the set of images to be fused rather than the HDR scene. They perform well if enough NER images are captured for the HDR scene. Unfortunately, this is nor always true. For example,  the brightness order could be reversed in the fused image if two LER images of an HDR scene are fused by them, and information in the brightest and darkest regions of an HDR scene might not be preserved well in the fused image if only a few NER images are fused for the HDR scene. All these problems will be addressed by the proposed ULMEF algorithm. With the proposed algorithm, the fused image approaches the HDR scene.

\section{The Proposed ULMEF Algorithm}
\label{paradigm}
The proposed ULMEF algorithm is inspired by the conventional MEF algorithms in \cite{1mertens2007,Li2017,kou2017} and the data-driven algorithms in \cite{1prabhakar2017,zheng2023efficient}, and it accepts any input sequence with arbitrary spatial resolution and exposure number. Novel unsupervised loss functions are proposed to train the proposed network, thereby avoiding the requirement of ground-truth images for the training. As such, the proposed ULMEF amgorithm is applicable to any set of differently exposed LDR images without camera movements and moving objects.

\begin{table*}[htb]
	\begin{center}
		\centering
		\caption{Definition of three different sets for an HDR scene}
		{\small \tabcolsep15pt\begin{tabular}{c|c|c}
				\hline 		
				set   & elements   & definition\\
				\hline
				$\Omega$ & $Z_1$, $Z_2$, $\cdots$, $Z_K$   & set of differently exposed images from an HDR scene \\
				$\Omega_f$ & $Z_{f(1)}$, $Z_{f(2)}$, $\cdots$, $Z_{f(\theta_1)}$   & set of differently exposed images to be fused \\
				$\Omega_m$ & $Z_{m(1)}$, $Z_{m(2)}$, $\cdots$, $Z_{m(\theta_2)}$   & set of differently exposed images to define loss functions \\
				\hline
		\end{tabular}}
		\label{differentset}
	\end{center}
\end{table*}

For simplicity, three sets are defined for an HDR scene as in table \ref{differentset}. The relationship among the three sets $\Omega$, $\Omega_f$ and $\Omega_m$ is
\begin{equation}
\Omega_f\subseteq \Omega_m\subseteq \Omega,
\end{equation}
$f(i)(1\leq i\leq \theta(1))$ and $m(i)(1\leq i\leq \theta(2))$ satisfy
\begin{align}
f(1)<f(2)<\cdots<f(\theta(1)),\\
m(1)<m(2)<\cdots<m(\theta(2)).
\end{align}

The exposure time of the LDR image $Z_k$ is denoted as $\Delta t_k$, and
\begin{equation}
\Delta t_1<\Delta t_2<\cdots <\Delta t_K.
\end{equation}

\subsection{Structure of the Proposed MSF-Net}
Given the set of differently exposed LDR images $\Omega_f$,  a multi-scale fusion network is designed to fuse them in a coarse-to-fine manner. As demonstrated in  Fig. \ref{Fig1}, a feature extract block is first adopted to transfer the input sequence from image domain to feature domain, adaptively extracting features that are beneficial for the fusion.  This is similar to the algorithms in \cite{1prabhakar2017, zheng2023efficient} in the sense that they are not based on the weight maps as the algorithms in \cite{MEFNet, jiang2023meflut}. The proposed algorithm and the algorithms in \cite{1prabhakar2017,zheng2023efficient} are good at avoiding halos from appearing in the fused images. A pyramid $\{F(\Omega_f)_{l=1}^L\}$ is then built from the full-resolution features $\{F(\Omega_f)\}$ by using the bi-cubic down-sampling with a scale factor of $L$.  Here, $F(\cdot)$ represents the feature extract block. The fused image is constructed at each scale by
\begin{align}
Z_F^l =  N(\upsilon^l)
\end{align}
where $N(\cdot)$ denotes the network for feature fusion module, and $\upsilon^l$ is computed as
\begin{align}
\upsilon^l  =  \left\{ \begin{array}{ll}
C(F(\Omega_f)^l) );&\mbox{if~} l=1\\
C(F(\Omega_f)^l, N_{-1}(F(\Omega_f)^{l-1}) \uparrow);&\mbox{otherwise}
\end{array} \right.,
\end{align}
$N_{-1}$ denotes the network $N$ without the last layer, $\uparrow$ is the up-sampling operation, and $C(\cdot)$ is the concatenation.

It is shown in Fig. \ref{Fig1} that the fused image with different scales can be constructed through the network $N$, and the multiple scales of the fused image are used to obtain the final image. The whole process is similar to the Gaussian pyramids and EPS pyramids in \cite{1mertens2007,Li2017,kou2017}, thus preserving the scene depth and local contrast of the fused image well.

The proposed MSF-Net is on top of the multi-scale recursive residual group (MSRRG) which has two attractive characteristics: (1) the structure of the MSRRG is a residual network \cite{residual}, it can reuse features to avoid the possible gradient vanishing, and (2) a novel multi-scale feature attention is used in  the MSRRG to suppress less useful features and only allow the propagation of more informative ones to effectively improve the quality of the fused image. As illustrated in Fig. \ref{Fig2}, the MSRRG mainly includes different scale convolutions and dual-attention blocks (DABs)  \cite{CycleISP}.  Each DAB combines a channel attention block and a spatial attention block in channel-wise and pixel-wise features, respectively. The DAB treats different features and pixels unequally, which can provide additional flexibility in dealing with different types of information. In addition, the proposed multi-scale structure is more beneficial for preserving fine details and scene depth for the fused image.

\begin{figure*}[htb!]
	\centering
	\includegraphics[width=1\textwidth]{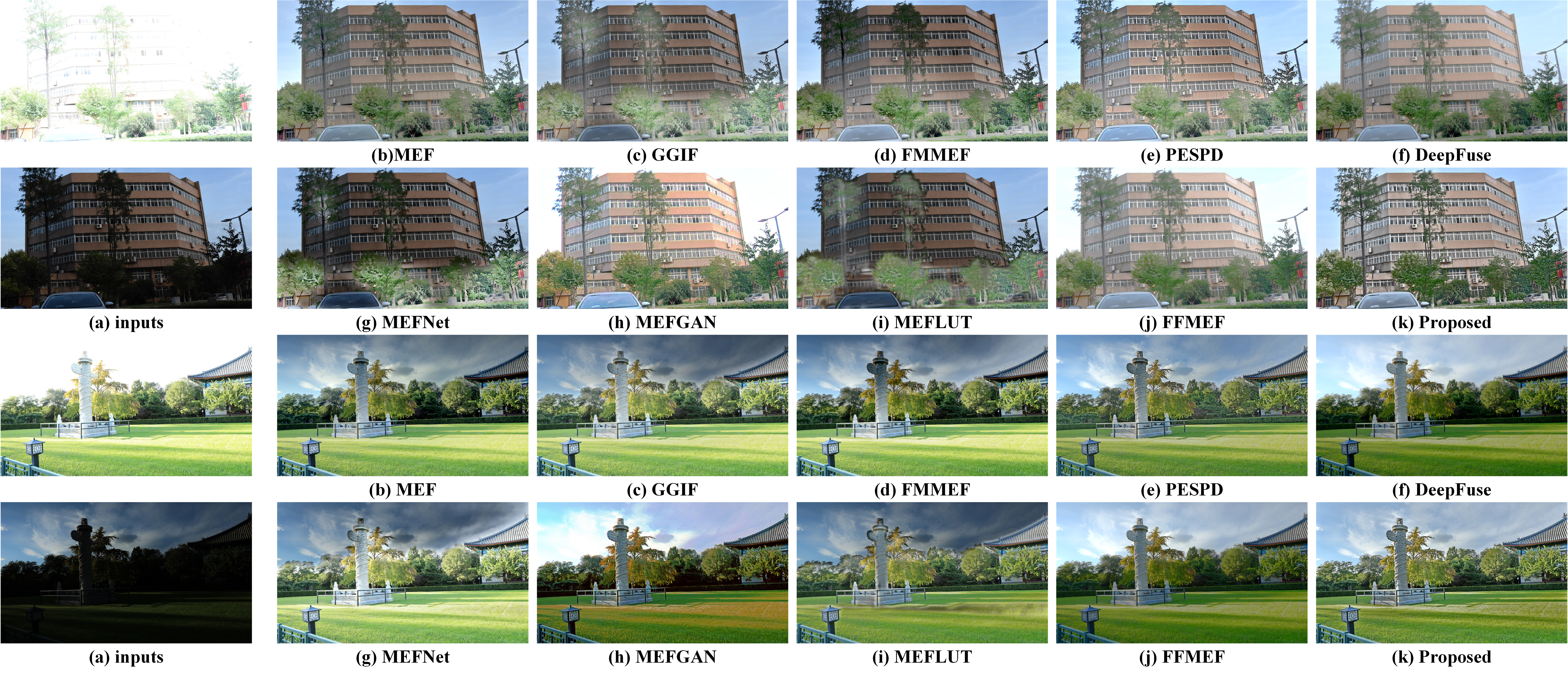}
	\caption{Visual comparison of ten different exposure fusion algorithms with the inputs as two LER images. The input data in the first column are from \cite{1zheng2023}. There are brightness order reversal artifacts in the fused images by the MEF \cite{1mertens2007}, GGIF \cite{kou2017}, FMMEF \cite{li2020fast}, PESPD \cite{zhang2023multi}, MEFNet \cite{MEFNet}, and MEFLUT \cite{jiang2023meflut}.}
	\label{Fig4}
\end{figure*}

\begin{figure*}[htb!]
	\centering
	\includegraphics[width=1\textwidth]{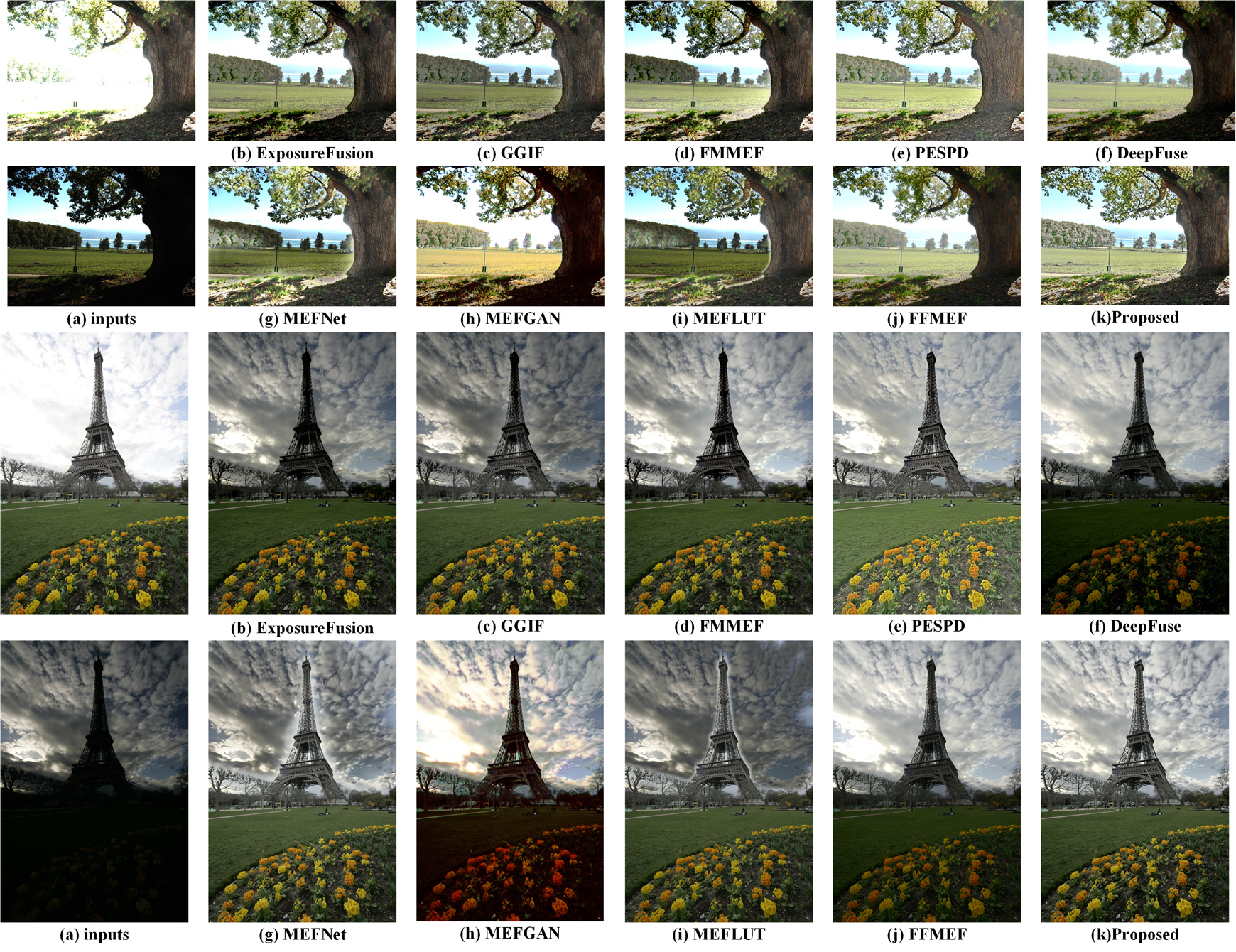}
	\caption{ Fusion results for comparison of different fusion algorithms with images. The input data in the first columns are from \cite{Cai2018}, the number of fused images is 2. It achieves stable fusion results across different domain datasets. There are halos in the fused images by the GGIF \cite{kou2017}, MEFNet \cite{MEFNet}, and MEFLUT \cite{jiang2023meflut}. Information in the brightest regions is not preserved well by the MEF \cite{1mertens2007}, PESPD \cite{zhang2023multi}, DeepFuse \cite{1prabhakar2017}, MEFGAN \cite{mefgan} and FFMEF \cite{zheng2023efficient}.}
	\label{Fig5}
\end{figure*}

\subsection{Unsupervised Loss Functions}

Besides the structure of the proposed MSF-Net, loss functions also play a crucial role for the proposed ULMEF algorithm. Since the ground-truth image is not available, unsupervised loss functions are defined to train the proposed MSF-Net.

Loss functions are defined by using the fused image $Z_F$ and the set of images to be fused $\Omega_f$ in the existing ULSEF algorithms \cite{1prabhakar2017, MEFNet, jiang2023meflut,zheng2023efficient}. The fused image approaches the set of images to be fused. Novel loss functions are proposed in this subsection by fully utilizing the asymmetry between the training and inferring (or testing) phases. Besides the set of images to be fused $\Omega_f$, other LDR images from the same HDR scene with different exposures are also used to define the loss functions. The fused image approaches the HDR scene. Clearly, the proposed loss functions and the set $\Omega_f$ are decoupled. The asymmetry between the training and inferring stages is well utilized by the proposed algorithm.

To preserve the scene contents in source images, the similarity constraint is implemented from two aspects: MEF-SSIM  quality measurement $L_{S}$ and weight mean absolute error (WAE) $L_{W}$. The overall loss function is thus defined as
\begin{equation}
\label{loss}
L(\Omega_m, Z_F) = L_{S}(\Omega_m, Z_F) + \lambda L_{W}(\Omega_m, Z_F),
\end{equation}
where $\lambda$ is a constant hyper-parameter to control the trade-off between the two aspects. The value of $\lambda$ is 10. Details on the $L_{S}(\Omega_m, Z_F)$ and $ L_{W}(\Omega_m, Z_F)$ are given in the appendix.

The loss function $L_S(\Omega_f, Z_F)$ is widely adopted in the ULSEF algorithms \cite{MEFNet, jiang2023meflut, 1prabhakar2017}, and it is defined by using the set of images to be fused $\Omega_f$. However, the proposed $L_{S}(\Omega_m, Z_F)$ is defined by using the set $\Omega_m$, and is fundamentally different from the $L_S(\Omega_f, Z_F)$ in the sense that the loss function and the set of images to be fused $\Omega_f$ are decoupled in the proposed $L_S(\Omega_m, Z_F)$. The loss function $L_W(\Omega_m, Z_F)$ is also new. Surprisingly, the new loss function $L_W(\Omega_m, Z_F)$ can improve the proposed ULMEF algorithm from the MEF-SSIM point of view.

\subsection{Exposure Interpolation and Exposure Extrapolation}

The proposed ULMEF is adopted to implement the exposure interpolation and exposure extrapolation as in the following two interesting cases:

{\bf Case 1} Exposure interpolation: the set $\Omega_f$ is a pair of two LER images $Z_{f(1)}$ and $Z_{f(2)}$ from an HDR scene \cite{1zheng2023}. The set $\Omega_m$ consists of the set $\Omega_f$ and the image with the middle exposure from the same HDR scene. $\theta_1$ is 2 and $\theta_2$ is 3. The objective of including the image with the middle exposure from the same HDR scene in the case 1 is to avoid the possible brightness order reversal from appearing in the fused image $Z_F$ \cite{1zheng2023}.

{\bf Case 2} Exposure extrapolation: the set $\Omega_f$ is a set of three NER images, $Z_{f(1)}$, $Z_{f(2)}$ and $Z_{f(3)}$ from an HDR scene. The set $\Omega_m$ includes the images in the set $\Omega_f$ and two more differently exposed images $Z_{f(1)-1}$ and $Z_{f(3)+1}$ from the same HDR scene. $\theta_1$ is 3 and $\theta_2$ is 5. The objective of including the images $Z_{f(1)-1}$ and $Z_{f(3)+1}$ in the case 2 is to further help preserve the information in the brightest and darkest regions of the HDR scene in the fused image $Z_F$.

\begin{figure*}[htb!]
	\centering
	\includegraphics[width=1\textwidth]{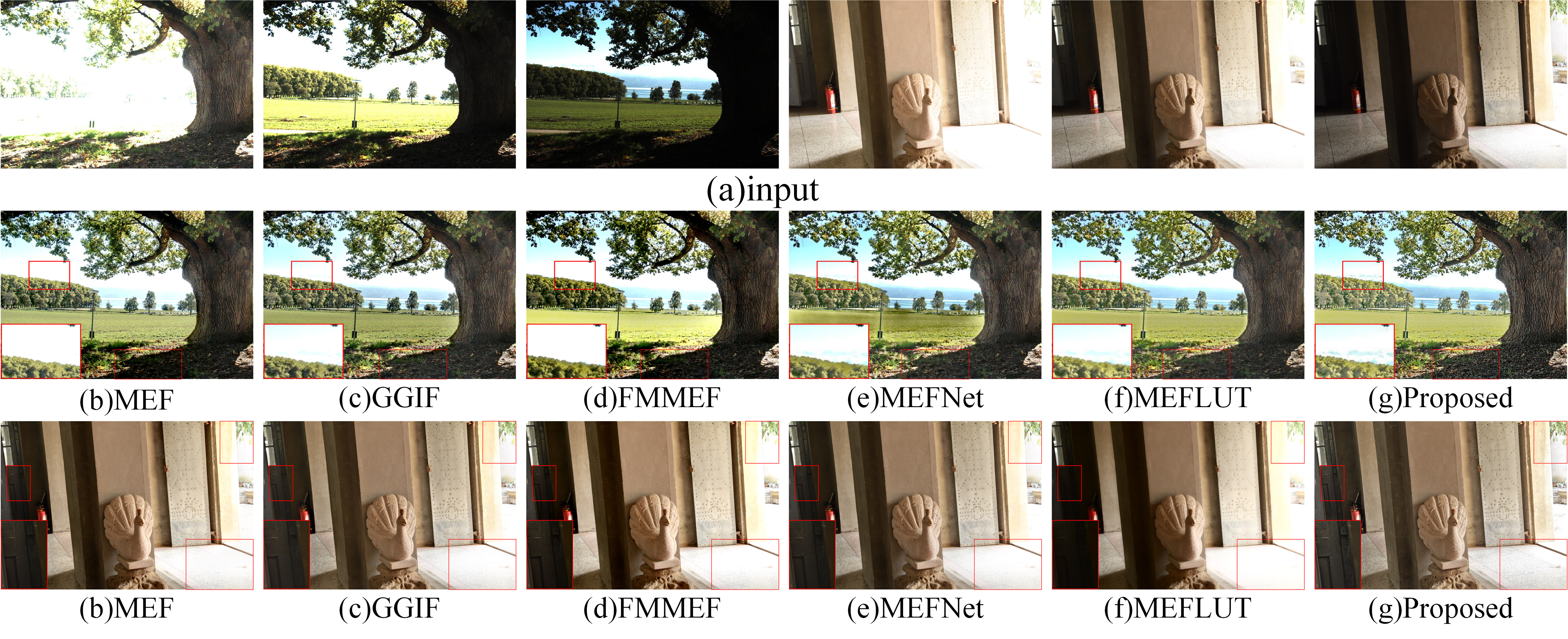}
	\caption{Comparison among the proposed algorithm and the algorithms in MEF \cite{1mertens2007}, GGIF \cite{kou2017},  FMMEF \cite{li2020fast}, MEFNet \cite{MEFNet}, MEFLUT \cite{jiang2023meflut} when the inputs are three NER images. As illustrated by the highlighted parts, information in the brightest and darkest regions of HDR scenes is much more visible regardless of display by the proposed algorithm.}
	\label{3imgs}
\end{figure*}

\section{Experimental Results}
\label{experimentalresult}
Experimental results are provided to validate the proposed ULMEF algorithm. Readers are invited to view to electronic version of figures and zoom in them so as to better check differences among all images. The dataset on HDR imaging in \cite{1zheng2023} is adopted to train and test all data-driven MEF algorithms. Camera shaking and object movement were strictly controlled to prevent them from appearing in the frame to capture static images \cite{1xuyl2021}.  The dataset is randomly split into three parts: 640 sequences for training, 50 ones for verifying, and the rest 100 ones for testing. To validate the generalization capability of different MEF algorithms, 50 sequences from the data set in \cite{Cai2018} were also used to test them.  We set the batch size to $1$. The learning rate is initially set to $10^{-4}$ and then decreased using a cosine annealing schedule for the training 200 epoches. All the experiments are implemented using PyTorch on NVIDIA A100.

\subsection{Comparison of Different MEF Algorithms}
The proposed ULMEF algorithm is first compared with four conventional exposure fusion  algorithms in \cite{1mertens2007, kou2017, li2020fast, zhang2023multi}  and five data-driven exposure fusion algorithms in \cite{1prabhakar2017, MEFNet, mefgan, jiang2023meflut, zheng2023efficient} in the case that the inputs are two LER images. The objective is to verify the efficiency of exposure interpolation.

\begin{table}[htb!]
	\begin{center}
		\caption{MEF-SSIM of ten different MEF Algorithms with two input images in the dataset \cite{1zheng2023} ($\uparrow$: larger is better)}
		{\small
			\tabcolsep5pt
			\scriptsize\begin{tabular}{ c|c|c|c|c}
				\hline
				MEF \cite{1mertens2007}  &  GGIF\cite{kou2017}  & FMMEF \cite{li2020fast}  & PESPD \cite{zhang2023multi}   & DeepFuse \cite{1prabhakar2017}   \\
				\hline
				0.9011	 &0.9035   & 0.9131  &0.9085     &0.9070   \\
				\hline
				MEFNet \cite{MEFNet}  &  MEFGAN \cite{mefgan}   & MEFLUT \cite{jiang2023meflut}   & FFMEF \cite{zheng2023efficient}   & Proposed   \\
				\hline
				0.8920	 &0.8499   & 0.8637  &0.8742     &0.9468   \\
				\hline	
			\end{tabular}
		}
		\label{tabfusion1}
	\end{center}
\end{table}

\begin{table}[htb!]
	\begin{center}
		\caption{MEF-SSIM of several different MEF Algorithms with two input images in the dataset \cite{Cai2018} ($\uparrow$: larger is better)}
		{\small
			\tabcolsep5pt
			\scriptsize\begin{tabular}{ c|c|c|c|c}
				\hline
				MEF \cite{1mertens2007}  &  GGIF \cite{kou2017}  & FMMEF \cite{li2020fast}  & PESPD \cite{zhang2023multi}   & DeepFuse \cite{1prabhakar2017}   \\
				\hline
				0.9357	    &0.9396     & 0.9408    &0.9282     &0.9072   \\
				\hline
				MEFNet \cite{MEFNet}  &  MEFGAN \cite{mefgan}   & MEFLUT \cite{jiang2023meflut}   & FFMEF \cite{zheng2023efficient}   & Proposed   \\
				\hline
				0.9237	 &0.6679   & 0.9135  &0.8965     &0.9452   \\
				\hline	
			\end{tabular}
		}
		\label{tabfusion2}
	\end{center}
\end{table}

All the ten algorithms are first compared from the subjective point of view. Particularly, they are compared from five points of view: halo artifacts, information in the brightest and darkest region, scene depth, local contrast, and brightness order reversal artifacts. As shown in Figs. \ref{Fig4} and \ref{Fig5}, the weight maps based algorithms in \cite{jiang2023meflut, kou2017, li2020fast, MEFNet, 1mertens2007, zhang2023multi} suffer from brightness order reversal artifacts, and the algorithms in \cite{jiang2023meflut, kou2017, MEFNet} suffer from halo artifacts even though the algorithms in \cite{jiang2023meflut, MEFNet} are much simpler than the algorithm in \cite{1prabhakar2017, zheng2023efficient}. As demonstrated in Figs. \ref{Fig4} and \ref{Fig5}, the modified arctan function in \cite{li2020fast, zhang2023multi} preserves the relative brightness order better than the Gaussian curve in \cite{kou2017, 1mertens2007}. Thus, the brightness order reversal artifacts are more serious in the fused images by the algorithms in \cite{li2020fast, zhang2023multi}. However, the Gaussian curve preserves the information in the brightest and darkest regions better. The single scale exposure fusion algorithm in \cite{jiang2023meflut, MEFNet, 1prabhakar2017} cannot preserve the scene depth and local contrast as the MEF algorithms in \cite{1mertens2007, zhang2023multi, kou2017, li2020fast}. Event though the learning algorithm in \cite{zheng2023efficient} is hierarchical, it cannot preserve the local contrast such as the grass in Fig. \ref{Fig4} well. The GAN based algorithm in \cite{mefgan} is on top of supervised learning and produces serious color distortions. The algorithms in \cite{1prabhakar2017, zheng2023efficient, mefgan} are not based on the weight maps. They are good at avoiding halo artifacts from appearing in the fused images however they cannot preserve the information in the darkest and brightest regions well. All these problems are overcome by the proposed ULMEF algorithm. Therefore, the exposure interpolation is important for HDR imaging on mobile devices with limited computational resource. Besides the subjective evaluation, all the ten algorithms are also compared from the MEF-SSIM point of view. The MEF- SSIM is calculated by using the fused image and the three captured images  which are the reference images. As shown in Tables \ref{tabfusion1} and \ref{tabfusion2},  the proposed algorithm achieves the highest MEF-SSIM especially for Table \ref{tabfusion2}, which is from the dataset \cite{Cai2018} and has noticeable domain differences. These results indicate that the proposed ULMEF  algorithm has strong generalization ability and is more robust to the dataset domain.

The proposed ULMEF algorithm is then compared with three conventional exposure fusion  algorithms in \cite{1mertens2007, kou2017, li2020fast}  and two data-driven exposure fusion algorithms in \cite{MEFNet, jiang2023meflut} in the case that the inputs are three differently exposed images. The objective is to verify the efficiency of exposure extrapolation. As shown in Figs. \ref{3imgs}, all the algorithms in \cite{1mertens2007,kou2017, MEFNet,jiang2023meflut, li2020fast} cannot preserve information in the brightest and darkest regions of the HDR scene well if the inputs are three NER images. This problem is overcome by the proposed ULMEF. Therefore, the exposure extrapolation is also very important for HDR imaging on mobile devices with limited computational resource.

\begin{figure*}[htb]
	\centering
	\includegraphics[width=1.0\textwidth]{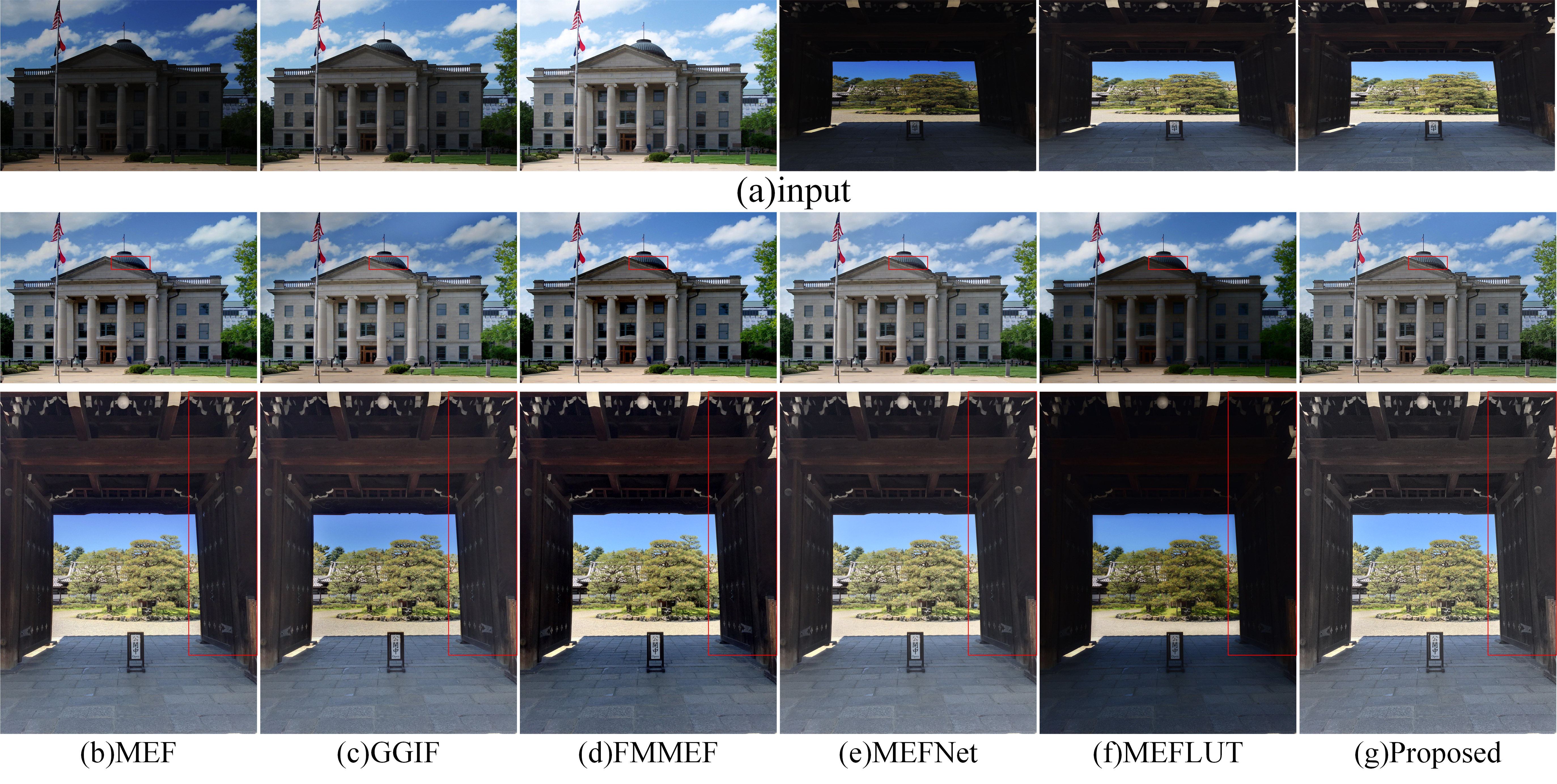}
	\caption{Fusion results for comparison of different fusion algorithms  in MEF \cite{1mertens2007}, GGIF \cite{kou2017},  FMMEF \cite{li2020fast}, MEFNet \cite{MEFNet}, MEFLUT \cite{jiang2023meflut} with images. The input data in the first columns are from \cite{Cai2018}, the number of fused images is 3.}
	\label{add}
\end{figure*}

More experiment results are provided to test the robustness of the proposed algorithm and the five SOTA algorithms including MEF \cite{1mertens2007}, GGIF \cite{kou2017},  FMMEF \cite{li2020fast}, MEFNet \cite{MEFNet}, and MEFLUT \cite{jiang2023meflut}. All the two sets of differently exposed images are from the data set in \cite{Cai2018}. Neither of the proposed algorithm and the algorithms \cite{MEFNet, jiang2023meflut} is trained by using the data from \cite{Cai2018}. As shown in Fig. \ref{add}, the proposed algorithm preserves information in the darkest and brightest regions of HDR scenes much better than the  in MEF \cite{1mertens2007}, GGIF \cite{kou2017},  FMMEF \cite{li2020fast}, MEFLUT \cite{jiang2023meflut}.

\subsection{Comparison of $L(\Omega_f, Z_F)$ and $L(\Omega_m, Z_F)$}
In this subsection, the conventional loss function $L(\Omega_f, Z_F)$ is compared with the proposed loss function $L(\Omega_m, Z_F$) by testing four sets of differently exposed images. There are two LER images $Z_{f(1)}$ and $Z_{f(2)}$ in each set $\Omega_f$. $\Delta t_{f(2)}$ is equal to $64\Delta t_{f(1)}$.  Each corresponding  set $\Omega_m$ includes the set $\Omega_f$ and one more image from the same HDR scene with the exposure time as $8\Delta t_{f(1)}$.
\begin{figure*}[htb!]
	\centering
	\includegraphics[width=1\textwidth]{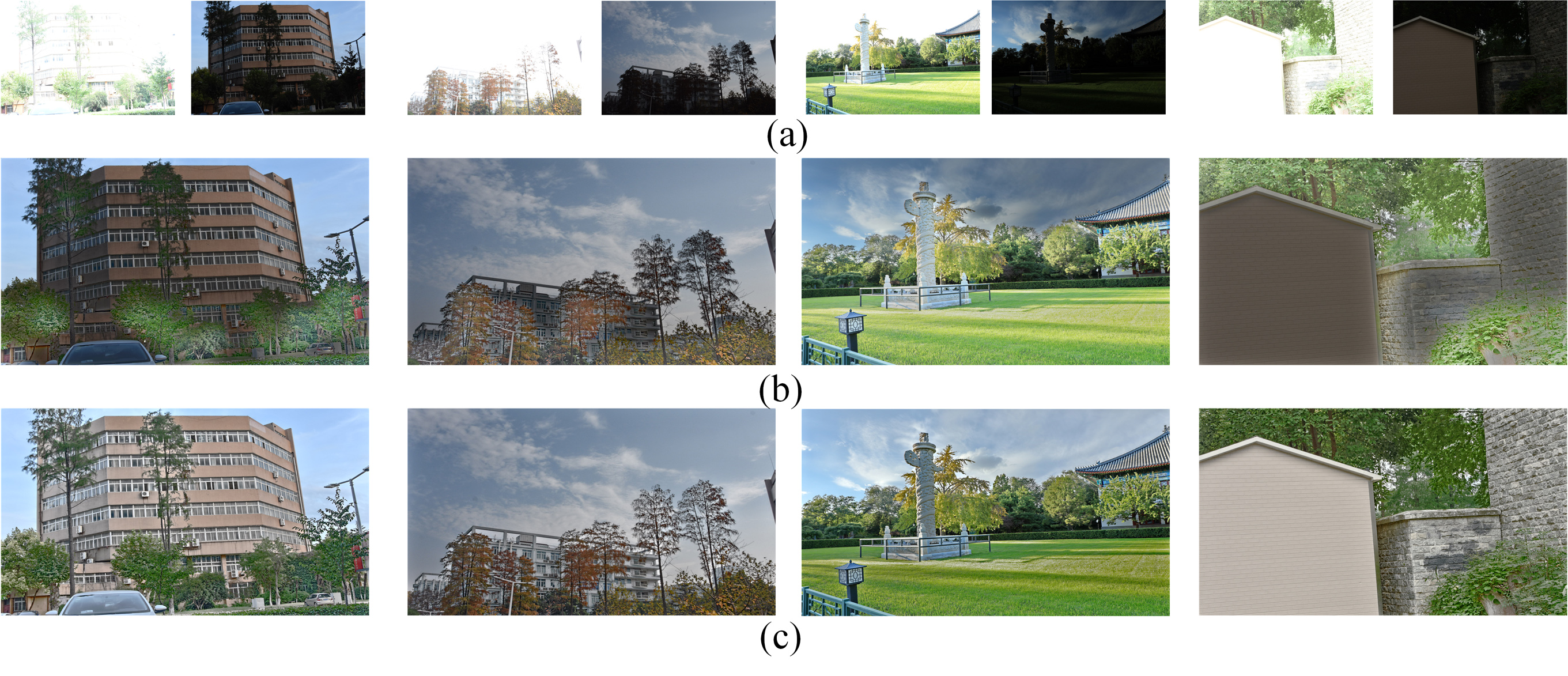}
	\caption{Comparison between the loss functions $L(\Omega_f, Z_F)$ and $L(\Omega_m, Z_F)$. (a) are the two inputs, (b) are fused images by $L(\Omega_f, Z_F)$ and (c) are the results of $L(\Omega_m, Z_F)$.  There are serious brightness order reversal artifacts in (b). The new initiative can significantly improve the quality of the fused images. }
	\label{Fig4a}
\end{figure*}

As shown in Fig. \ref{Fig4a}, there are serious brightness order reversal artifacts in all the four images fused by using the loss function  $L(\Omega_f, Z_F)$. For example, the trees are darker than the sky in the first three images and the wall in the fourth image in the inputs but they are brighter than the sky and wall in the fused images. The problem is overcome by the proposed loss function  $L(\Omega_m, Z_F)$, because each fused image approaches each HDR scene with the guidance from the loss function $L(\Omega_m, Z_f)$. Clearly, the proposed loss function significantly outperforms the loss function  $L(\Omega_f, Z_F)$.

\subsection{Ablation Study of Two Other Key Components}
Two other key components of the proposed framework are: 1) multi-scale, and 2) loss function  $L_W$. Their
performances are evaluated in this subsection.

\begin{figure*}[htb]
	\centering
	\includegraphics[width=1\textwidth]{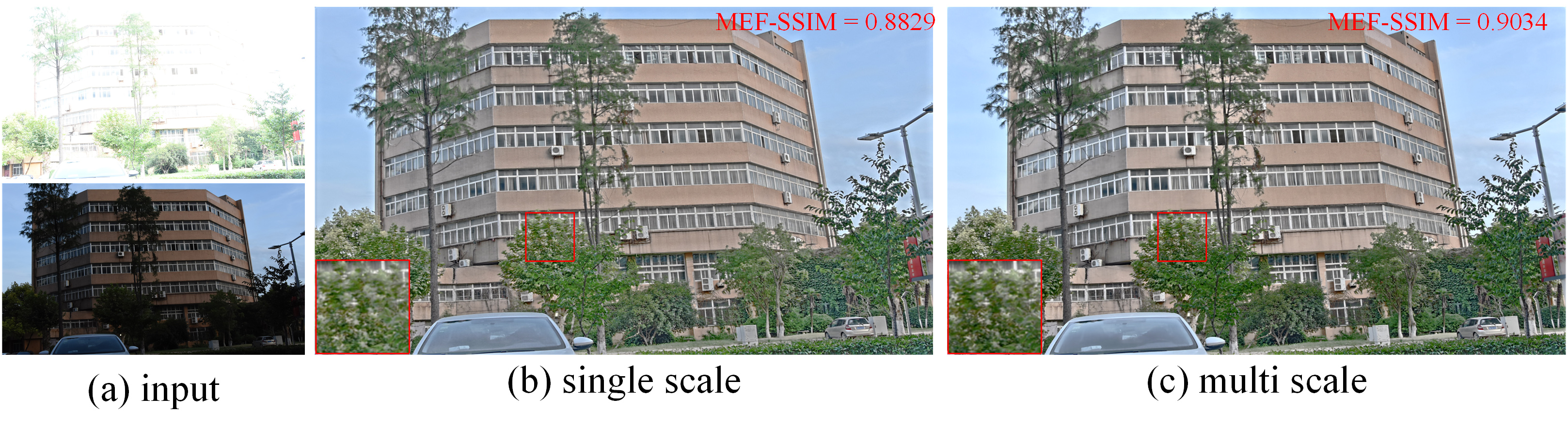}
	\caption{Comparison between the single-scale and multi-scale. (a) are the two inputs,  (b) is a fused image by the single scale and (c) is a fused image by the multi-scale. Both the scene depth and local contrast are preserved better in (c).}
	\label{Fig6a}
\end{figure*}

There are two multi-scale components in the proposed MSF-Net: the hierarchical structure and the MSSRG. Neither of them is enabled when the multi-scale is disabled. As shown in Table \ref{Ablation}, there is noticeable gain from the MEF-SSIM point of view by using the multi-scale. Meanwhile, it can be shown from the zoom-in region in Fig. \ref{Fig6a} that both the scene depth and local contrast are indeed preserved better by using the multi-scale components.

The proposed $L(\Omega_m, Z_F)$ includes two components. The component $L_S$ is widely used in the existing unsupervised learning based exposure fusion algorithms. The new component $L_W$ improves the MEF-SSIM as demonstrated in Table \ref{Ablation}.

\begin{table}[htb]
	\begin{center}
		\centering
		\caption{Ablation study on two more key components of the proposed ULMEF on \cite{1zheng2023}  ($\uparrow$: larger is better)}
		{\small \tabcolsep15pt\begin{tabular}{c|cc|c}
				\hline 		
				Case   & Multi-Scale   & $L_W$   & MEF-SSIM ($\uparrow$)  \\
				\hline
				1 & Y   & N      & 0.9460 \\
				2 & N   & Y      & 0.9452 \\
				3 & Y   & Y      & 0.9468 \\
				\hline
		\end{tabular}}
		\label{Ablation}
	\end{center}
\end{table}

\subsection{Limitation of The Proposed Method}
Same as the DeepFuse \cite{1prabhakar2017}, the complexity of the proposed method would be an issue if it was applied to fuse differently exposed images with a large size. It is interesting to combine the GRU, multi-scale, and the proposed loss functions to develop a new MEF algorithm. This method will be investigated in our future research.

\section{Conclusion Remarks}
\label{conclusion}
A novel unsupervised learning based multi-scale exposure fusion (ULMEF) algorithm is proposed for merging a set of different exposed low dynamic range (LDR) images into a high-quality LDR image for a high dynamic range (HDR) scene. The fused image approaches the HDR scene rather than the set of LDR images to be fused. Therefore, the proposed algorithm can avoid halo and brightness order reversal artifacts from appearing in the fused image and preserve the scene depth and local contrast as well as information in the darkest and brightest regions well in the fused image. In addition,  experimental results show that the proposed algorithm can produce better fusion images than several state-of-the-art exposure fusion algorithms when only a few differently exposed LDR images are fused for an HDR scene. The proposed algorithm well utilizes the asymmetry between the training and inferencing (or testing) stages of the learning based algorithm and the conventional wisdom of inferring  better via seeing more. This is a new initiative on the exposure fusion. We believe that better exposure fusion algorithms would be developed along the initiative in future.

\section*{Appendix: The Proposed Loss Functions}

Details on the proposed WAE and SSIM-MEF are provided in this appendix.

{\it WAE}: The loss function $L_{W}$ is used to constrain the intensity distribution differences of images at the pixel level. Inspired by the conventional MEF algorithms in \cite{1mertens2007,Li2017,kou2017}, a  weight function is used to measure reliable information from all the differently exposed images in the set $\Omega_m$. This is different from the MEF algorithms in [10, 12,16] in the sense that  the WAE is defined by the set $\Omega_f$ in [10,12,16]. The weight function $\bar{W}_k(p)$ is obtained by considering contrast $C$, saturation $S$ and well-exposedness $E$, and it is first computed as $C_k(p) \times S_k(p) \times E_k(p)$, and then normalized by the values of the $\theta(2)$ weight maps such that they sum to one at each pixel $p$, i.e.
\begin{equation}
W_{m(k)}(p) = \frac{\bar{W}_{m(k)}(p)}{\sum_{k'=1}^{\theta_2} \bar{W}_{m(k')}(p)}.
\end{equation}

In order to reduce sharp weight map transitions, the normalized weight maps are smoothed by using the iWGIF \cite{1jia2022} with the guidance image as the luminance channel of each image. More reliable areas containing bright colors and details will be assigned larger weights, so that the network will pay more attention to obtain more reliable information. The loss function $L_W(\Omega_m, Z_F)$ is defined as
\begin{equation}
\label{loss1}
L_{W}(\Omega_m, Z_F) = \sum_{k'=1}^{\theta(2)} \sum_p W_{m(k')}(p)\|Z_F(p)-Z_{m(k')}(p)\|_1.
\end{equation}

{\it MEF-SSIM}: Since the MEF-SSIM index in \cite{1Ma2015} is effective to measure the quality of the fused image, it is also selected as an objective function. Similarly, the MEF-SSIM is defined by using the fused image $Z_F$ and the set $\Omega_m$.

Let $R_i(\cdot)$ is an operator that extracts the $i$-th patch from an image, i.e., $R_i(Z_{m(k')})$ is the $i$th patch extracted from the image $Z_{m(k')}$ in the set $\Omega_m$. The MEF-SSIM index decomposes $R_i(Z_{m(k')})$ into three conceptually independent components as
\begin{align}
R_i(Z_{m(k')}) = c_{m(k'),i}s_{m(k'),i}+ l_{m(k'),i},
\end{align}
where $l_{m(k'),i}$, $c_{m(k'),i}$, and $s_{m(k'),i}$ represent the intensity, contrast, and structure of the patch $R_i(Z_{m(k')})$ respectively as
\begin{align}
l_{m(k'),i} &= \mu_{R_i(Z_{m(k')})},\\
c_{m(k'),i}&=\|R_i(Z_{m(k')}) - \mu_{R_i(Z_{m(k')})} \|_2,\\
s_{m(k'),i}&= \frac{R_i(Z_{m(k')}) - \mu_{R_i(Z_{m(k')})}}{ \|R_i(Z_{m(k')}) - \mu_{R_i(Z_{m(k')})} \|_2},
\end{align}
and $\mu_{R_i(Z_{m(k')})}$ is  the mean intensity of the patch $R_i(Z_{m(k')})$.

Since a higher contrast means a patch with a higher quality, the desired contrast is determined by the highest contrast by using all the images in the set  $\Omega_m$ as
\begin{equation}
\label{loss3}
\hat{c}_i = \max_{1\leq k' \leq \theta(2)}\{c_{m(k'),i}\},
\end{equation}
and the desired structure is also defined by using all the images in the set  $\Omega_m$ as
\begin{align}
\label{loss4}
\hat{s}_i & =  \frac{\bar{s}_i}{\|\bar{s}_i\|_2}, \\
\overline{s}_i & = \frac{\sum_{k'=1}^{\theta(2)} \|R_i(Z_{m(k')}) - \mu_{R_i(Z_{m(k')})}\|_{\infty} s_{m(k'),i}} {\sum_{k'=1}^{\theta(2)} \| R_i(Z_{m(k')}) - \mu_{R_i(Z_{m(k')})}\|_{\infty}}.
\end{align}

The desired  intensity of the fused patch is computed by a weighted summation over all the images in the set  $\Omega_m$ as
\begin{equation}
\label{loss5}
\widehat{l}_i= \frac{\sum_{k'=1}^{\theta(2)} w_l(\mu_{m(k')}, l_{m(k'),i}) l_{m(k'),i}} {\sum_{k'=1}^{\theta(2)} w_l(\mu_{m(k')}, l_{m(k'),i})},
\end{equation}
where $w_l(\cdot)$ is defined by using the global mean intensity $\mu_{m(k')}$ of the image $Z_{m(k')}$ and the local mean intensity $l_{m(k'),i}$ of the patch $R_i(Z_{m(k')})$ as
\begin{equation}
\label{loss6}
w_l(\mu_{m(k')}, l_{m(k'),i}) =  \exp(-\frac{(\mu_{m(k')}-\tau)^2} {2\sigma_g ^2}-\frac{(l_{m(k'),i}-\tau)^2} {2\sigma_l ^2}),
\end{equation}
$\sigma_g$ and $\sigma_l$ are set as $0.2$ and $0.5$, respectively, and $\tau$ is $0.5$.

The desired fused patch $R_i(\widehat{Z})$ is then computed by
\begin{align}
R_i(\widehat{Z})  = \widehat{c}_i  \widehat{s}_i+ \widehat{l}_i,
\end{align}
and the  MEF-SSIM index of the patches $R_i(\Omega_{f})$ is  defined as
\begin{align}
\nonumber
S(R_i(\Omega_{m}), R_i(Z_F)) = &\frac{2\mu_{R_i(\widehat{Z})} \mu_{ R_i(Z_F)} + C_1}{\mu_{R_i(Z_F)}^2+ \mu_{R_i(\widehat{Z})}^2 + C_1}\\\label{loss7}
&\frac{2\sigma_{R_i(\widehat{Z})R_i(Z_F)} + C_2 }{\sigma_{R_i(Z_F)}^2+ \sigma_{ R_i(\widehat{Z})}^2 + C_2},
\end{align}
where $\sigma_{R_i(Z_F)}^2$ is the variances of the patch $R_i(Z_F)$, $\sigma_{R_i(\widehat{Z})R_i(Z_F)}$ is the covariance between the patches $R_i(\widehat{Z})$ and $R_i(Z_F)$. $C_1$ and $C_2$ are two small positive constants to prevent the possible instability.

The MEF-SSIM loss function $L_{S}(\Omega_m, Z_F)$ is finally defined as
\begin{align}
\label{loss8}
L_{S}(\Omega_m, Z_F) = 1 -\frac{1}{M}\sum_{i=1}^M S(R_i(\Omega_m), R_i(Z_F)),
\end{align}
where $M$ is the number of blocks.

%\bibliographystyle{unsrt}
%\bibliography{main}

\end{document}